\documentclass{article} 
\usepackage{conference,times}
\usepackage{algorithm}
\usepackage{listings}
\usepackage{graphicx}
\usepackage{amsmath}
\usepackage{amssymb}
\usepackage{multirow}
\usepackage{subfigure}
\usepackage{makecell}
\usepackage{boldline}
\usepackage{algorithm}
\usepackage{listings}
\usepackage{bbding}
\iclrfinalcopy

\usepackage{amsmath,amsfonts,bm}

\def\eqref#1{equation~\ref{#1}}

\def\1{\bm{1}}

\DeclareMathAlphabet{\mathsfit}{\encodingdefault}{\sfdefault}{m}{sl}
\SetMathAlphabet{\mathsfit}{bold}{\encodingdefault}{\sfdefault}{bx}{n}

\usepackage{hyperref}
\hypersetup{
colorlinks=true,
    citecolor = {blue}
}

\usepackage{url}

\newcommand{\ma}[1]{\mathbf{#1}}
\newcommand{\te}[1]{\bm{\mathcal{#1}}}

\title{\centering S$^2$-MLPv2: Improved Spatial-Shift MLP  Architecture for Vision}

\author{\hspace{1.2in}Tan Yu, Xu Li, Yunfeng Cai,  Mingming Sun,  Ping Li \\\\
\hspace{1.9in}Cognitive Computing Lab\\
\hspace{2.15in}Baidu Research\\
\hspace{1.2in}10900 NE 8th St. Bellevue, Washington 98004, USA\\
\hspace{1.22in}No.10 Xibeiwang East Road, Beijing 100193, China\\
\hspace{0.75in}\{tanyu01,\ lixu13,\ caiyunfeng,\ sunmingming01,\ liping11\}@baidu.com \\
}

\begin{document}

\maketitle

\begin{abstract}

Recently,  MLP-based vision backbones emerge. MLP-based vision architectures with less inductive bias achieve competitive performance in image recognition compared with CNNs and vision Transformers.  Among them, spatial-shift MLP (S$^2$-MLP), adopting the straightforward spatial-shift operation, achieves better performance than the pioneering works including  MLP-mixer and ResMLP.
More recently, using smaller patches with a pyramid  structure, Vision Permutator (ViP)  and Global Filter Network (GFNet) achieve better performance than  S$^2$-MLP.
 In this paper, we improve the S$^2$-MLP vision backbone. We expand the feature map along the channel dimension and split the  expanded feature map into several parts. We conduct different spatial-shift operations on split parts.
 Meanwhile, we exploit the split-attention operation to fuse these split parts. Moreover, like the counterparts, we adopt smaller-scale patches and use a pyramid structure for boosting the image recognition accuracy.   We term the improved  spatial-shift MLP vision backbone as S$^2$-MLPv2.  Using 55M parameters, our medium-scale model, S$^2$-MLPv2-Medium achieves  an $83.6\%$ top-1 accuracy on the ImageNet-1K benchmark using $224\times 224$ images without self-attention and external training data.

\end{abstract}

\section{Introduction}

Recently,  extensive studies  on computer vision are conducted to achieve high performance with less inductive bias.  Two types of  architectures emerge including vision Transformers~\citep{dosovitskiy2020image,touvron2020training}  and  MLP-based backbones~\citep{tolstikhin2021mlp,touvron2021resmlp}.  Compared with \emph{de facto} vision backbone CNN~\citep{he2016deep} with delicately devised  convolution kernels,  both  vision Transformers  and  MLP-based backbones have achieved competitive performance in image recognition without expensive hand-crafted  design.  Specifically, vision Transformer models  stack a series of Transformer blocks, achieving the global reception field.

 MLP-based methods such as MLP-Mixer~\citep{tolstikhin2021mlp} and ResMLP~\citep{touvron2021resmlp} achieve the  communication between patches  through projections along different patches implemented by MLP.   Different from  MLP-Mixer and ResMLP, spatial-shift MLP (S$^2$-MLP)~\citep{yu2021s}  adopts a very straightforward operation, spatial shifting, for communications between patches, achieving higher image recognition accuracy on ImageNet1K dataset without external training data.  In parallel,  Vision Permutator (ViP)~\citep{hou2021vision} encodes the feature representation along the height and width dimensions and meanwhile exploits the finer patch size with a two-level pyramid  structure, achieving better performance than   S$^2$-MLP.
 CCS-MLP~\citep{yu2021rethinking} devises a circulant token-mixing MLP for achieving the translation-invariance property.
 Global Filter Networks  (GFNet)~\citep{rao2021global} exploits 2D Fourier Transform to map the spatial patch features into  the frequency domain and conducts the  cross-patch communications in the frequency domain.  As pointed by~\cite{rao2021global}, the token-mixing operation in the frequency domain is equivalent to depthwise convolution with circulant weights.   To achieve a high recognition accuracy,  GFNet also utilizes   patches of smaller size with a pyramid structure. More recently, AS-MLP~\citep{lian2021mlp}   axially shifts channels of the feature map and devises a four-level pyramid, achieving excellent performance. In parallel,  Cycle-MLP~\citep{chen2021cyclemlp} devises several pseudo-kernels for spatial projection and also achieves outstanding performance. It is worth noting that, both AS-MLP~\citep{lian2021mlp} and Cycle-MLP~\citep{chen2021cyclemlp} are based on  the well-devised four-level pyramid.

\begin{figure}
    \centering
     \subfigure[Spatial-shift operations in S$^2$-MLP]{
        \includegraphics[scale=0.48]{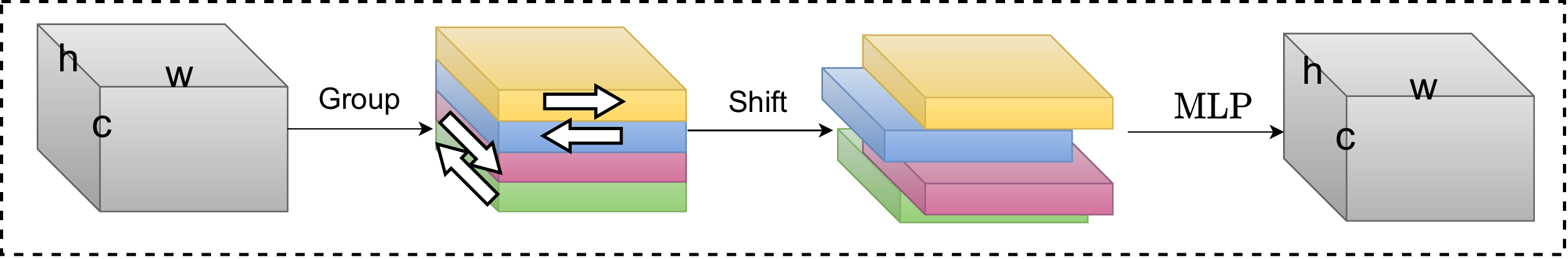}}
    \subfigure[Spatial-shift operations in S$^2$-MLPv2]{
        \includegraphics[scale=0.48]{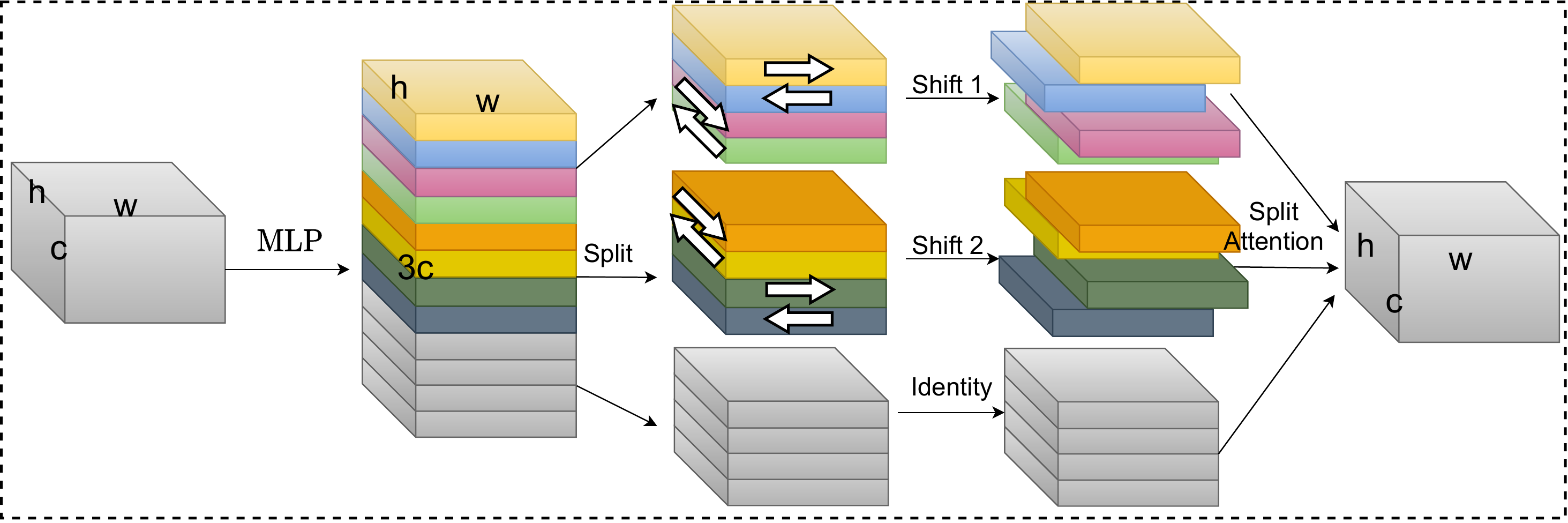}}
   \caption{Comparisons between the spatial-shift operations in S$^2$-MLP~\citep{yu2021s} and the proposed  S$^2$-MLPv2. In S$^2$-MLP, the channels are equally divided into four parts, and each part shifts along different directions. An MLP is conducted on the shifted channels. In contrast, in  S$^2$-MLPv2, the $c$-channel feature map is  expanded into the
   $3c$-channel feature map. Then the  expanded map is equally split into three parts along the channel dimension. For each part, we conduct different spatial-shift operations. Then the shifted parts are merged through the split-attention operation~\citep{zhang2020resnest} to generate the $c$-channel feature map.}
    \label{visual}\vspace{0.2in}
\end{figure}

In this work, we rethink the design of spatial-shift MLP (S$^2$-MLP)~\citep{yu2021s} and propose an improved spatial-shift MLP (S$^2$-MLPv2). Compared with the original S$^2$-MLP, the modifications are mainly conducted on two aspects:
\begin{itemize}
\item As visualized in Figure~\ref{visual} (b),  we expand the feature map along the channel dimension and split the expanded feature map into multiple parts. For different parts, we conduct different spatial-shift operations.  We  exploit  the split-attention operation ~\citep{zhang2020resnest}  to fuse these split parts.
\item We adopt smaller-scale patches and the hierarchical pyramid structure  like existing MLP-based architectures such as  ViP~\citep{hou2021vision}, GFNet~\citep{rao2021global}, AS-MLP~\citep{lian2021mlp} and Cycle-MLP~\citep{chen2021cyclemlp}.
\end{itemize}

We term the improved  spatial-shift MLP  architecture as S$^2$-MLPv2.  We visualize the difference between the original spatial-shift MLP (S$^2$-MLP) and the improved verision, S$^2$-MLPv2, in Figure~\ref{visual}.
Our experiments conduct on the public benchmark, ImageNet-1K, demonstrates the state-of-the-art image recognition accuracy of the proposed S$^2$-MLPv2. Specifically, using 55M parameters, our medium-scale model, S$^2$-MLPv2-Medium achieves $83.6\%$ top-1 accuracy using $224\times 224$ images without self-attention and external training data. 

\newpage

\section{Related Work}

\textbf{vision Transformer.} vision Transformer (ViT)~\citep{dosovitskiy2020image} crops an image into $16\times 16$ patches, and treat each patch as a token in the input of Transformer.  These patches/tokens are processed by a stack of Transformer layers  for communications with each other.  It has achieved competitive  image recognition accuracy as CNNs using huge-scale pre-training datasets. DeiT~\citep{touvron2020training} adopts more advanced optimizer as well as data augmentation methods, achieving promising results using medium-scale pre-training datasets.   Pyramid vision transformer (PvT)~\citep{wang2021pyramid} and PiT~\citep{heo2021pit} exploit a pyramid structure which gradually shrinks the spatial dimension and expands the hidden size, achieving better performance.
Tokens-to-Token (T2T)~\citep{yuan2021tokens} and Transformer-in-Transformer (TNT)~\citep{han2021transformer} improve the effectiveness of  modeling the local structure of each patch/token.
To overcome the inefficiency of the global self-attention,  Swin~\citep{liu2021swin} conducts the self-attention within local windows but achieves the global reception field through shifting the window settings.  Shuffle Transformer~\citep{huang2021shuffle} also exploits the local self-attention windows and achieves the cross-window communications through switching the spatial dimension and the feature dimension. Twins~\citep{chu2021twins} enhances the  self-attention within local windows by the global sub-sampled attention.  DynamicViT~\citep{rao2021dynamicvit} and SViTE~\citep{chen2021chasing} exploit the sparsity for achieving high efficiency.
CaiT~\citep{touvron2021going} explores the extremely deep architecture  by stacking tens of layers.  Recently,  PVTv2~\citep{wang2021pvtv2} improves PvT using  overlapping patch embedding, convolutional feedforward networks, and linear-complexity attention layers.
CSwin Transformer~\citep{dong2021cswin} improves Swin through cross-shaped windows computing self-attention in the horizontal and vertical stripes in parallel.
 Focal Transformer~\citep{yang2021focal} also develops more advanced local windows which attend fine-grain tokens only locally, but the summarized ones globally.

\vspace{0.05in}

\textbf{MLP-based architectures.} MLP-mixer~\citep{tolstikhin2021mlp} is the pioneering work for MLP-based vision backbone. It proposes a token-mixing MLP consisting of two fully-connected layers for communications between patches.  Res-MLP~\citep{touvron2021resmlp} simplifies the  token-mixing MLP to a single fully-connected layer and explores the deeper architecture with more layers.
Spatial-shift MLP backbone (S$^2$-MLP)~\citep{yu2021s} adopts the spatial-shift operation for cross-patch communications.  Vision Permutator (ViP)~\citep{hou2021vision} mixes tokens along the height dimension and the width dimension, separately. Meanwhile, ViP adopts a pyramid structure as   PvT~\citep{wang2021pyramid} and achieves considerably better performance than MLP-mixer, Res-MLP and S$^2$-MLP. CCS-MLP~\citep{yu2021rethinking} rethinks the design of token-mixing MLP in  MLP-mixer and Res-MLP, and propose a circulant channel-specific MLP. Specifically, they devise the weight matrix of token-mixing MLP as a circulant matrix, taking fewer parameters.  Meanwhile, the multiplication between  vector and circulant matrix can be efficiently computed through Fast Fourier  Transform (FFT). Global Filter Network (GFNet)~\citep{rao2021global} maps the patch features to the frequency domain through 2D FFT and mixes the tokens in the frequency domain. As proved by~\cite{rao2021global}, the global filter in GFNet is equivalent to  a depthwise global circular convolution with the filter size H $\times$ W.  Meanwhile,  GFNet also exploits pyramid structure for boosting the recognition accuracy.  In this work, we rethink the design of S$^2$-MLP and considerably improves its performance in image recognition.

\section{Preliminary}

\subsection{Spatial-shift MLP (S$^2$-MLP)}
In this section, we briefly review the structure of S$^2$-MLP~\citep{yu2021s} architecture. It consists of the patch embedding layer,  a stack of S$^2$-MLP blocks and the classification head.

\noindent \textbf{Patch embedding layer.} It first crops an image of $W\times H \times 3$ size into $w\times h$ patches.  Each patch is of $p\times p \times 3$ size and $p = \frac{W}{w} = \frac{H}{h}$.
It then  maps each patch into a $d$-dimensional vector through a fully-connected layer.

\begin{figure}[htp!]
\centering
\includegraphics[scale=0.95]{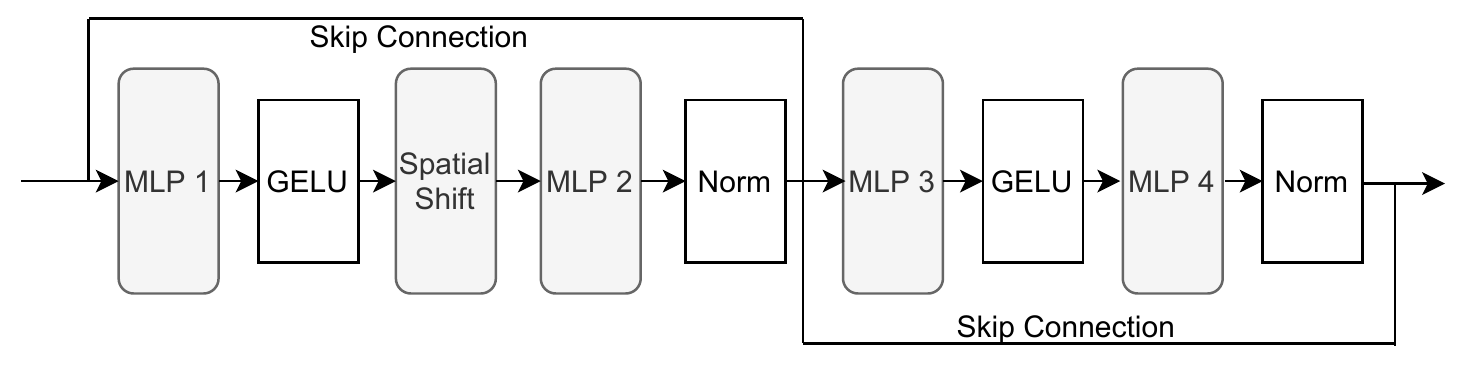}
\caption{The structure of an S$^2$-MLP block.}
\label{ssmlp}
\end{figure}

\noindent \textbf{Spatial-shift MLP block.} As visualized in Figure~\ref{ssmlp},  it consists of four MLP layers for mixing channels  and a spatial shift layer for mixing patches.  Below we only introduce the spatial-shift module. Given an input tensor $\mathcal{X} \in \mathbb{R}^{w\times h \times c}$, it first equally splits $\te{X}$ into four parts $\{\te{X}_i\}_{i=1}^4$ along the channel dimension and  shifts them along four directions:
\begin{equation}
\label{eq:ss}
\begin{split}
&\te{X} [2:h,:,1:c/4] \gets \te{X}[1:h-1,:,1:c/4],\\
&\te{X}[1:h-1,:,c/4+1:c/2] \gets \te{X}[2:h,:,c/4+1:c/2], \\
& \te{X}[:,2:w,c/2:3c/4] \gets \te{X}[:,1:w-1,c/2:3c/4],\\
&\te{X}[:,1:w-1,3c/4:c] \gets \te{X}[:,2:w,3c/4:c].
\end{split}
\end{equation}

It is worth noting that,  S$^2$-MLP~\citep{yu2021s} stacks $N$   Spatial-shift MLP blocks with the same settings and does not exploit pyramid structure as its MLP-backone counterparts such as  Vision Permutator~\citep{hou2021vision} and Global Filter Network (GFNet)~\citep{rao2021global}.

\subsection{Split Attention}
Vision Permutator~\citep{hou2021vision} adopts split attention proposed in ResNeSt~\citep{zhang2020resnest} for enhancing multiple feature maps from different operations. Specifically, we denote $K$ features maps of the same size $n\times c$  by  $[\ma{X}_1, \ma{X}_2, \cdots, \ma{X}_K]$ where $n$ is the number of patches and $c$ is the number of channels,  the split-attention operation first averages them and obtains
\begin{equation}
\label{split1}
\ma{a} =   \sum_{k=1}^K \mathbf{1} \ma{X}_k,
\end{equation}
where $\mathbf{1}  \in \mathbb{R}^{n}$ is the $n$-dimensional row vector with all $1$s.
Then $\ma{a} \in \mathbb{R}^c$ goes through a stack of MLPs  and generates
\begin{equation}
\label{split2}
\hat{\ma{a}} =\sigma(\ma{a} \ma{W}_1) \ma{W}_2,
\end{equation}
where $\sigma$ is the activation function implemented by GELU,  $\ma{W}_1 \in \mathbb{R}^{c\times \bar{c}}$ and $\ma{W}_2 \in \mathbb{R}^{\bar{c}\times Kc}$ are weights of MLPs and the output $\hat{\ma{a}}  \in \mathbb{R}^{Kc}$.  Then  $\hat{\ma{A}}$ is reshaped into a matrix $\hat{\ma{{A}}}  \in \mathbb{R}^{ K \times c}$, which is further processed by a softmax function along the first dimension and generates $ \bar{\ma{{A}}} = \mathrm{softmax}(\hat{\ma{A}}) \in \mathbb{R}^{K \times c}$.
Then it generates the attended feature map $\hat{\ma{X}}$ where each row of $\hat{\ma{X}}$, $\hat{\ma{X}}[i,:]$, is computed by

\begin{equation}
\label{split3}
\hat{\ma{X}}[i,:] = \sum_{k=1}^K \ma{X}_k[i,:] \odot \bar{\ma{A}}[k,:],
\end{equation}

where $\odot$ denotes the element-wise multiplication between two vectors.

\section{ S$^2$-MLPv2}

In this section, we introduce the proposed S$^2$-MLPv2 architecture. Similar to S$^2$-MLP backbone, S$^2$-MLPv2 backbone  consists of the patch embedding layer,  a stack of S$^2$-MLPv2  blocks and the classification head.  Since we have introduced the patch embedding layer in the previous section, we only introduce the proposed S$^2$-MLPv2  block below.

\subsection{S$^2$v2 block}
\begin{figure}[htp!]
\centering
\includegraphics[scale=0.85]{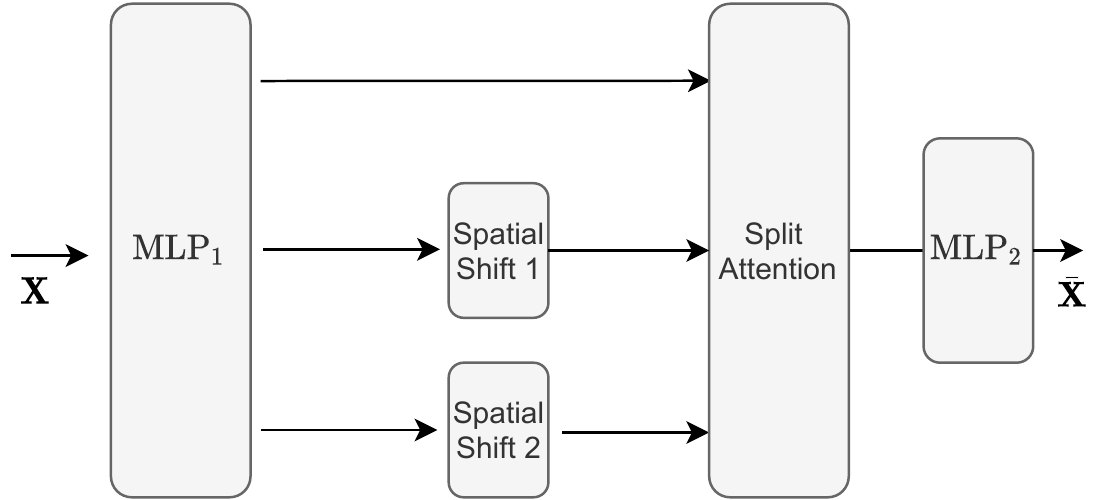}
\caption{The structure of an S$^2$-MLPv2 block.}
\label{v2module}
\end{figure}
The S$^2$-MLPv2 block consists of two parts, the S$^2$-MLPv2 component and the channel-mixing  MLP (CM-MLP) component.
Given an input feature map $\te{X} \in \mathbb{R}^{w \times h \times c}$, it conducts
\begin{equation}
\begin{split}
\te{Y} &= \mathrm{S}^2\textit{-}\mathrm{MLPv2}(\mathrm{LN}(\te{X})) + \te{X},\\
\te{Z} &= \mathrm{CM}\textit{-}\mathrm{MLP}(\mathrm{LN}(\te{Y})) + \te{Y}.\\
\end{split}
\end{equation}
The channel-mixing MLP (CM-MLP) adopts the same structure as MLP-mixer \citep{tolstikhin2021mlp} and ResMLP~\citep{touvron2021resmlp} and thus we skip their details here. Below we only introduce the proposed  S$^2$-MLPv2 component  in detail.

Given an input feature map $\te{X} \in \mathbb{R}^{w \times h \times c}$, the proposed  S$^2$-MLPv2 component  first expands the channels of $\te{X}$ from $c$ to $3c$ by an MLP:
\begin{equation}
\hat{\te{X}} = \mathrm{MLP}_1(\te{X}) \in \mathbb{R}^{w \times h \times 3c}.
\end{equation}
Then it equally splits the expanded feature map $\hat{\te{X}}$ along the channel dimension into three parts:
\begin{equation}
{\te{X}}_1 = \hat{\te{X}}[:,:,1:c], {\te{X}}_2 = \hat{\te{X}}[:,:,c+1:2c],   {\te{X}}_3 = \hat{\te{X}}[:,:,2c+1:3c].
\end{equation}
It shifts ${\te{X}}_1$ and ${\te{X}}_2$ through two  two spatial-shift layers $\mathrm{SS}_{1}(\cdot)$ and  $\mathrm{SS}_{2}(\cdot)$.   $\mathrm{SS}_{1}(\cdot)$  conducts the same  spatial-shift operation as \eqref{eq:ss}.  In contrast, $\mathrm{SS}_{2}(\cdot)$ conducts an asymmetric spatial-shift operation with respect to $\mathrm{SS}_{1}(\cdot)$. To be specific,  given the feature map $\te{X}_2$, $\mathrm{SS}_{2}(\te{X}_2)$ conducts:
\begin{equation}
\label{eq:ss2}
\begin{split}
&\te{X}_2 [:,2:w,1:c/4] \gets \te{X}_2[:,1:w-1,1:c/4],\\
&\te{X}_2[:,1:w-1,c/4+1:c/2] \gets \te{X}_2[:,2:w,c/4+1:c/2], \\
& \te{X}_2[2:h,:,c/2:3c/4] \gets \te{X}_2[1:h-1,:,c/2:3c/4],\\
&\te{X}_2[1:h-1,:,3c/4:c] \gets \te{X}_2[2:h,:,3c/4:c].
\end{split}
\end{equation}

\begin{algorithm}[t]
\caption{Pseudocode of our  S$^2$-MLPv2 module.}
\label{alg:code}
\lstset{
  backgroundcolor=\color{white},
  basicstyle=\fontsize{9.5pt}{9.5pt}\ttfamily\selectfont,
  columns=fullflexible,
  breaklines=true,
  captionpos=b,
  commentstyle=\fontsize{9.5pt}{9.5pt}\color{codeblue},
  keywordstyle=\fontsize{9.5pt}{9.5pt},
}
\begin{lstlisting}[language=Python]

def spatial_shift1(x):
     b,w,h,c = x.size()
     x[:,1:,:,:c/4] = x[:,:w-1,:,:c/4]
     x[:,:w-1,:,c/4:c/2] = x[:,1:,:,c/4:c/2]
     x[:,:,1:,c/2:c*3/4] = x[:,:,:h-1,c/2:c*3/4]
     x[:,:,:h-1,3*c/4:] = x[:,:,1:,3*c/4:]
     return x

def spatial_shift2(x):
     b,w,h,c = x.size()
     x[:,:,1:,:c/4] = x[:,:,:h-1,:c/4]
     x[:,:,:h-1,c/4:c/2] = x[:,:,1:,c/4:c/2]
     x[:,1:,:,c/2:c*3/4] = x[:,:w-1,:,c/2:c*3/4]
     x[:,:w-1,:,3*c/4:] = x[:,1:,:,3*c/4:]
     return x

class S2-MLPv2(nn.Module):
    def __init__(self, channels):
        super().__init__()
        self.mlp1 = nn.Linear(channels,channels*3)
        self.mlp2 = nn.Linear(channels,channels)
        self.split_attention = SplitAttention()
    def forward(self, x):
        b,w,h,c = x.size()
        x = self.mlp1(x)
        x1 = spatial_shift1(x[:,:,:,:c/3])
        x2 = spatial_shift2(x[:,:,:,c/3:c/3*2])
        x3 = x[:,:,:,c/3*2:]
        a = self.split_attention(x1,x2,x3)
        x = self.mlp2(a)
        return x
\end{lstlisting}
\end{algorithm}

Note that we intentionally devise $\mathrm{SS}_{1}(\cdot)$ and  $\mathrm{SS}_{2}(\cdot)$ in an asymmetric structure so that they are complementary to each other.
Meanwhile, we do not shift ${\te{X}}_3$ and just keep it as well.

After that,   $\{\te{X}_k\}_{k=1}^3$ are reshape into matrices $\{\ma{X}_k\}_{k=1}^3$ where $\ma{X}_k \in \mathbb{R}^{wh\times c}$, which are fed into a split-attention (SA) module as \eqref{split1}, \eqref{split2}, and \eqref{split3},  to generate
\begin{equation}
\label{sa}
\hat{\ma{X}} = \mathrm{SA}(\{{\mathbf{X}}_k\}_{k=1}^3).
\end{equation}
Then the attended feature map $\mathbf{A}$  is  further fed into another MLP layer for generating the output
\begin{equation}
\bar{\ma{X}} =  \mathrm{MLP}_2(\hat{\mathbf{X}}).
\end{equation}
The structure of the proposed S$^2$-MLP module is visualized in Figure~\ref{v2module}  and the details are listed in Algorithm~\ref{alg:code}.

\subsection{Pyramid structure}
Following vision Permutator~\citep{hou2021vision}, we also exploit the two-level pyramid structure to enhance the performance.
To make a fair comparison with Vision Permutator~\citep{hou2021vision}, we adopt the exact same pyramid structure. The details are in Table~\ref{pdetails}.
We notice that counterpart works such as PVTv2~\citep{wang2021pvtv2},  AS-MLP~\citep{lian2021mlp}, and Cycle-MLP~\citep{chen2021cyclemlp} adopt more advanced pyramid structure with smaller patches in the early blocks. The smaller patches might be better at capturing the fine-grained visual details and lead to higher recognition accuracy.  Nevertheless, due to the limited computing resources, it is unfeasible for us to re-implement  all these pyramid structures.    Moreover, we also notice that Vision Permutator devises a large model with considerably more parameters and FLOPs. Nevertheless, due to the limited computing resources, the large model is  not feasible for us, either.

\vspace{0.15in}

\begin{table}[htp!]
\begin{tabular}{c|cccc|cccc|c}
 \hlineB{3}
Settings & \makecell{Patch \\ Size} & \makecell{$\#$ of \\ Tokens} &  \makecell{Hidden\\ Size} &  \makecell{$\#$ of \\ Blocks} &  \makecell{Patch\\  Size} & \makecell{$\#$ of \\ Tokens} &  \makecell{Hidden \\ Size} &   \makecell{$\#$ of \\ Blocks} &  \makecell{Expa. \\ Ratio} \\ \hlineB{3}
Small/7       &      $7\times 7$        &   $32^2$      &      $192$        &  $4$ &  $2\times 2$          &  $16^2$       &     $384$        &  $14$ &      $3$           \\
Medium/7      &       $7 \times 7 $     &  $32^2$      &     $256$        &  $7$ &    $2\times 2$           & $16^2$       &      $512$       & $17$ &     $3$            \\  \hlineB{3}
\end{tabular}
\caption{The configurations of the two-level pyramid structure used in  our S$^2$-MLPv2. We exploit both small and medium settings, which are the exactly same as Vision Permutator~\citep{hou2021vision}  fo a  fair comparison. AS-MLP~\citep{lian2021mlp} and Cycle-MLP~\citep{chen2021cyclemlp} adopt more advanced four-level pyramid structure with patches of the smaller scale. }
\label{pdetails}
\end{table}

\newpage

\section{Experiments}
We testify the proposed S$^2$-MLPv2 on ImageNet-1K dataset~\citep{deng2009imagenet}. We do not use external data for training. The implementation is based on the PaddlePaddle deep learning platform.

\vspace{0.05in}

\textbf{Implementation details.}
Following DeiT~\citep{touvron2020training}, we adopt AdamW~\citep{loshchilov2018decoupled} as optimizer.  We train both the small model and the medium model using four NVIDIA A100 GPU cards. For the small model, we set the batch size as 1024.
In contrast, for the medium model, we only set the batch size as 744  due to the GPU memory limitation  of four NVIDIA A100 GPU cards.  We set the initial learning rate as 2e-3 and decay it to 1e-5 in 300 epochs using a cosine function. The weight decay rate is set to be 5e-2 following previous works~\citep{touvron2020training,hou2021vision}.  We also conduct warming up in the first 10 epochs following~\cite{hou2021vision}.  Moreover, as adopted by~\cite{touvron2020training,hou2021vision}, we conduct multiple data augmentation methods including Rand-Augment~\citep{cubuk2020randaugment}, Mixup~\citep{zhang2017mixup} and CutMix~\citep{yun2019cutmix} and CutOut~\citep{zhong2020random}.  Like DeiT~\citep{touvron2020training} and Vision Permutator~\citep{hou2021vision}, we adopt exponential
moving average (EMA) model~\citep{laine2016temporal}.
Meanwhile, we also use label smoothing~\citep{szegedy2016rethinking} with a smooth ratio of $0.1$
and DropPath~\citep{huang2016deep} with a drop ratio of $0.1$ for both small  and medium settings.

\subsection{Comparisons with state-of-the-art methods}

\begin{table}[htp!]
\centering
\begin{tabular}{c|cccccc}
\hlineB{3}
Model   & Pyramid  & Para. & FLOPs & \makecell{Train \\ Size} & \makecell{Test \\ Size} & \makecell{Top-1 \\ Acc. ($\%$)} \\ \hlineB{3}
\multicolumn{7}{c}{Small models}  \\ \hlineB{3}

EAMLP-14~\citep{guo2021beyond}  &  &   30M        & $-$      &    224      &      224     &       $78.9$     \\
ResMLP-S24~\citep{touvron2021resmlp} & & 30M           &  6.0B      &      224      & 224          & $79.4$            \\
gMLP-S~\citep{liu2021pay}   && 20M          &  4.5B     &      224      &  224         & $79.6$  \\
GFNet-S~\citep{rao2021global} &  &  25M	   &4.5B & 224&224 & $80.0$ \\
GFNet-H-S~\citep{rao2021global} & \checkmark &  32M	   &4.5B & 224&224 & $81.5$ \\
AS-MLP-T~\citep{lian2021mlp} & \checkmark & 28M   & 4.4B & 224 & 224 & $81.3$ \\
CycleMLP-B2~\citep{chen2021cyclemlp} & \checkmark &   27M&  3.9B & 224&224 & $81.6$ \\
ViP-Small/7~\citep{hou2021vision} & \checkmark  & 25M&  6.9B & 224&224 & $81.5$ \\
\textbf{S$^2$-MLPv2-Small/7 (ours)} & \checkmark & 25M   &  6.9B & 224&224 & $\mathbf{82.0}$     \\\hlineB{3}
\multicolumn{7}{c}{Medium models}  \\ \hlineB{3}
MLP-mixer~\citep{tolstikhin2021mlp} & &    59M        & 11.6B      &    224      &      224     &       $76.4$ \\
EAMLP-19~\citep{guo2021beyond}  &  &  55M        & $-$      &    224      &      224     &       $79.4$     \\
S$^2$-MLP-deep~\citep{yu2021s}  &  & 51M & 10.5B & 224 & 224 & $80.7$ \\
CCS-MLP-36~\citep{yu2021rethinking}&   & 43M & 8.9B & 224 & 224 & $80.6$ \\
GFNet-B~\citep{rao2021global} &  &  43M	   &7.9G & 224&224 & $80.7$ \\
GFNet-H-B~\citep{rao2021global} &  \checkmark &  54M &	8.4B & 224&224 & $82.9$ \\
AS-MLP-S~\citep{lian2021mlp} &   \checkmark& 50M&  8.5B & 224 & 224 & $83.1$ \\
CycleMLP-B4~\citep{chen2021cyclemlp} &   \checkmark & 52M&  10.1B & 224&224 & $83.0$ \\
ViP-Medium/7~\citep{hou2021vision} & \checkmark  & 55M& 16.3B  & 224&224 & $82.7$ \\
\textbf{S$^2$-MLPv2-Medium/7~(ours)} &  \checkmark &55M& 16.3B  & 224&224 & $\mathbf{83.6}$ \\ \hlineB{3}
\multicolumn{7}{c}{Large models}  \\ \hlineB{3}
CycleMLP-B5~\citep{chen2021cyclemlp} &   \checkmark & 76M&  12.3B & 224&224 & $83.2$ \\
AS-MLP-B~\citep{lian2021mlp} &   \checkmark& 88M&  15.2B & 224 & 224 & $83.3$ \\
ViP-Large/7~\citep{hou2021vision} & \checkmark  & 88M& 24.3B  & 224&224 & $83.2$ \\ \hlineB{3}
\end{tabular}
\caption{Comparisons with  MLP-like backbones on ImageNet-1K benchmark without extra data. Our S$^2$-MLPv2-Medium/7 achieves the state-of-the-art performance on the benchmark among medium-scale MLP models and even outperforms the existing large-scale MLP models. M denotes million and B denotes billion.}
\label{mlps}
\end{table}

\textbf{Comparisons with existing MLP-like methods.} In Table~\ref{mlps}, we compare our  S$^2$-MLPv2 with existing MLP-like backbones including MLP-Mixer~\citep{tolstikhin2021mlp}, EAMLP~\citep{guo2021beyond},   ResMLP~\citep{touvron2021resmlp}, gMLP~\citep{liu2021pay},
S2-MLP-deep~\citep{yu2021s}, CCS-MLP~\citep{yu2021rethinking},  GFNet~\citep{rao2021global},  AS-MLP~\citep{lian2021mlp}, CycleMLP~\citep{chen2021cyclemlp} and ViP~\citep{hou2021vision} on both small and base settings.   Among them, MLP-Mixer, ResMLP, gMLP,  S2-MLP, CCS-MLP do not exploit the pyramid structure,  and thus they cannot achieve competitive recognition accuracy compared with GFNet,  AS-MLP, CycleMLP, and ViP, which are with  the pyramid structure as shown in Table~\ref{mlps}. Meanwhile, as shown in the table, our  S$^2$-MLPv2 consistently outperforms its counterparts in both small and medium settings using a comparable number of parameters. Meanwhile, our medium model performs even better than the large models of AS-MLP~\citep{lian2021mlp}, CycleMLP~\citep{chen2021cyclemlp} and ViP~\citep{hou2021vision} with considerably more parameters. On the other hand, we notice that  both S$^2$-MLPv2 and ViP take more FLOPs compared with GFNet, AS-MLP,  CycleMLP. This is due to that ours and ViP use a very coarse pyramid structure, whereas   GFNet, AS-MLP,  CycleMLP use a more advanced pyramid. Both S$^2$-MLPv2 and ViP might potentially reduce FLOPs by using a  well-devised pyramid structure like GFNet, AS-MLP,  CycleMLP.

\vspace{0.05in}

\textbf{Comparisons with CNNs and vision Transformers.}
We compare the proposed  S$^2$-MLPv2  with  CNN models including  ResNet50~\citep{he2016deep}, RegNet~\citep{radosavovic2020designing} and  EfficientNet~\citep{tan2019efficientnet}. Meanwhile, we also compare with   vision Transformer models including ViT~\citep{dosovitskiy2020image}, DeiT~\citep{touvron2020training}, PVT~\citep{wang2021pyramid}, T2T~\citep{yuan2021tokens}, TNT~\citep{han2021transformer}, PiT~\citep{heo2021pit}, MViT~\citep{fan2021multiscale} CaiT~\citep{touvron2021going}, Swin~\citep{liu2021swin},    Shuffle Transformer~\citep{huang2021shuffle}, Nest Transformer~\citep{zhang2021aggregating},    Focal Transformer~\citep{yang2021focal}, and CSWin~\citep{dong2021cswin}.

\vspace{0.05in}

As shown in Table~\ref{main}, our  S$^2$-MLPv2-Medium achieves comparable accuracy as its vision Transformer counterparts using fewer parameters but more FLOPs. Using a more advanced pyramid as PvTv2, the FLOPs of our  S$^2$-MLPv2-Medium might be  reduced.
Noting that, compared with vision Transformer requiring complex self-attention operations, ours is much simpler in formulation and takes consideraly fewer parameters, making it a competitive choice in practical deployment.
\vspace{0.1in}

\begin{table*}[htpb!]
\centering
\begin{tabular}{c|cccc}
\hlineB{3}
Model & Scale & Top-1 ($\%$)  & Params (M) & FLOPs (B) \\
\hlineB{3}
\multicolumn{5}{c}{CNN-based models}  \\ \hlineB{3}
ResNet50~\citep{he2016deep}  &  $224 $    &  $76.2$    &    $25.6$ & $4.1$   \\
RegNetY-16GF~\citep{radosavovic2020designing}  &  $224 $    &  $80.4$       &    $83.6$ & $15.9$   \\
EfficientNet-B3~\citep{tan2019efficientnet}  &  $300 $    &  $81.6$       &    $12$ & $1.8$
\\
EfficientNet-B5~\citep{tan2019efficientnet}  &  $456$    &  $84.0$      &    $30$ & $9.9$
\\
\hlineB{3}
\multicolumn{5}{c}{Transformer-based models}  \\ \hlineB{3}
ViT-B/16$^{*}$~\citep{dosovitskiy2020image}&  $224$    &  $79.7$        &    $86.4$ & $17.6$  \\
DeiT-B/16~\citep{touvron2020training} &   $224 $   &  $81.8$          &  $86.4$ & $17.6$     \\
PVT-L~\citep{wang2021pyramid} &    $224$   &  $82.3$           &  $61.4$ & $9.8$     \\
TNT-B~\citep{han2021transformer} &    $224 $   &  $82.8$          &  $65.6$ & $14.1$     \\
T2T-24~\citep{yuan2021tokens} &    $224$   &  $82.6$       &  $65.1$ & $15.0$     \\
CPVT-B~\citep{chu2021conditional} &    $224 $   &  $82.3$      &  $88$ & $17.6$     \\
PiT-B/16~\citep{heo2021pit} &   $224$   &  $82.0$       &  $73.8$ & $12.5$     \\
MViT-B-24~\citep{fan2021multiscale} & $224$   &  $83.1$       &  $53.5$ & $10.9$     \\
CaiT-S32~\citep{touvron2021going} &    $224 $   &  $83.3$      &  $68$ & $13.9$     \\
Swin-B~\citep{liu2021swin} &    $224 $   &  $83.3$         &  $88$ & $15.4$    \\
Shuffle-B~\citep{huang2021shuffle} &    $224 $   &  $84.0$         &  $88$ & $15.6$    \\
Nest-B~\citep{zhang2021aggregating} &    $224$   &  $83.8$          &  $68$ & $17.9$ \\
PvTv2-B4~\citep{wang2021pvtv2}              &    $224$   &  $83.6$          &  $62.6$ & $10.1$ \\
Focal-Base~\citep{yang2021focal} & $224$   &  $83.8$          &  $89.8$ & $16.0$ \\
CSWin-B~\citep{dong2021cswin}              &    $224$   &  $84.2$          &  $78$ & $15.0$
\\ \hlineB{3}
\multicolumn{5}{c}{Our models}  \\ \hlineB{3}
S$^2$-MLPv2-Small/7 &    $224$   &  $82.0$       &  $25$ & $6.9$    \\
S$^2$-MLPv2-Medium/7 &    $224$   &  $83.6$    &  $55$ & $16.3$     \\ \hlineB{3}
\end{tabular}
\caption{Comparisons with CNN and Transformer models on ImageNet-1K  without extra data.  ViT-B/16$^{*}$ denotes the result of ViT-B/16 reported by~\cite{tolstikhin2021mlp} with extra regularization. Compared with Transformer-based models, our S$^2$-MLPv2-Medium/7 model achieves comparable recognition accuracy on the benchmark without self-attention and considerably fewer parameters.}
\label{main}
\end{table*}

\subsection{Ablation studies}

\textbf{Influence of the pyramid structure.}
To evaluate the influence of the pyramid structure on the proposed S$^2$-MLPv2, we compare the Small/7 settings  and  the Small/14  settings. The details of Small/7 settings  and  the Small/14  settings are in Table~\ref{pyramid}.  Both of them are the same as that in Vision Permutator~\citep{hou2021vision}.
Specifically, the initial patch size in Small/7  is $7\times 7$, which is smaller than the $14\times 14$ patches in  the Small/14 settings.  Intuitively, the smaller patches are beneficial to modeling fine-grained details in the images  and tend to achieve higher recognition accuracy. Table~\ref{pyresult} compares the performance of these two settings. As shown in the table, by utilizing the pyramid,  S$^2$-MLPv2-Small/7   achieves a considerably  better performance than S$^2$-MLPv2-Small/14.

\begin{table}[htp!]
\centering
\begin{tabular}{c|cccc|cccc|c}
 \hlineB{3}
Settings & \makecell{Patch \\ Size} & \makecell{$\#$ of \\ Tokens} &  \makecell{Hidden\\ Size} &  \makecell{$\#$ of \\ Blocks} &  \makecell{Patch\\  Size} & \makecell{$\#$ of \\ Tokens} &  \makecell{Hidden \\ Size} &   \makecell{$\#$ of \\ Blocks} &  \makecell{Expa. \\ Ratio} \\ \hlineB{3}
Small/7       &      $7\times 7$        &   $32^2$      &      $192$        &  $4$ &  $2\times 2$          &  $16^2$       &     $384$        &  $14$ &      $3$           \\
Small/14      &       $14 \times 14 $     &  $16^2$      &     $384$        &  $4$ &    $2\times 2$           & $16^2$       &      $384$       & $14$ &     $3$            \\  \hlineB{3}
\end{tabular}
\caption{The configurations of the  Small/7 settings with the pyramid structure and  Small/14 without  the pyramid structure.}
\label{pyramid}\vspace{0.1in}
\end{table}

\begin{table}[htp!]
\centering
\begin{tabular}{c|cccc}
 \hlineB{3}
Settings & Pyramid  & \makecell{Top-1 ($\%$)} & \makecell{$\#$ of parameters} &  \makecell{FLOPs} \\ \hlineB{3}
S$^2$-MLPv2-Small/7       &  \checkmark &    $82.0$        &   $25$M      &      $6.9$B             \\
S$^2$-MLPv2-Small/14      &    &   $80.9$     &  $30$M      &     $5.7$B                 \\  \hlineB{3}
\end{tabular}
\caption{Comparisons between Small/7  and Small/14 settings.}
\label{pyresult}
\end{table}

\vspace{0.15in}

\textbf{Influence of the split attention.} Recall from \eqref{sa} that,   we use the split-attention (SA) for fusing the feature maps  $\mathbf{A} = \mathrm{SA}( \{{\mathbf{X}}_k\}_{k=1}^3)$ . An alternating fusing manner is sum-pooling them  implemented by $\mathbf{A} = {\sum_{k=1}^3 {\mathbf{X}}_k }/3$. We compare  these two manners in Table~\ref{sasum}. The experiments are conducted in Small/7 settings. As shown in Table~\ref{sasum}, the split attention significantly outperforms sum pooling with a  slight increase in the number of parameters and FLOPs.

\vspace{0.1in}

\begin{table}[htp!]
\centering
\begin{tabular}{c|ccc}
 \hlineB{3}
Settings   & \makecell{Top-1 ($\%$)} & \makecell{$\#$ of parameters} &  \makecell{FLOPs} \\ \hlineB{3}
Sum-pooling       &    $79.8$        &   $22$M      &      $6.9$B             \\
Split-attention     &       $82.0$     &  $25$M      &     $6.9$B                 \\  \hlineB{3}
\end{tabular}
\caption{Performance comparisons between split attention  and sum pooling. The experiments are conducted in Small/7 settings.}
\label{sasum}
\end{table}

\vspace{0.05in}

\textbf{Influence of each split.}  As \eqref{sa},  we fuse three splits  $\{{\mathbf{X}}_k \}_{k=1}^3$ through the split attention. In this section, we evaluate the influence of removing one of them.
The experiments are conducted in Small/7 settings. As shown in Table \ref{split},  when using only ${\mathbf{X}}_1$ and ${\mathbf{X}}_3$, the top-1 accuracy drops from $82.0\%$  to $81.6\%$. Meanwhile, when removing ${\mathbf{X}}_3$, the top-1 accuracy decreases to $81.6\%$.

\vspace{0.1in}

\begin{table}[htp!]
\centering
\begin{tabular}{c c c|ccc}
 \hlineB{3}
 ${\mathbf{X}}_1$ &  ${\mathbf{X}}_2$ & ${\mathbf{X}}_3$   & \makecell{Top-1 ($\%$)} & \makecell{$\#$ of parameters} &  \makecell{FLOPs} \\ \hlineB{3}
\checkmark &  \checkmark  &  \checkmark &       $82.0$        &   $25$M      &      $6.9$B             \\
\checkmark &   & \checkmark &           $81.6$     &  $22$M      &     $6.2$B                 \\
\checkmark &  \checkmark &  &           $81.7$     &  $22$M      &     $6.2$B                 \\  \hlineB{3}
\end{tabular}
\caption{The influence of each split. The experiments are conducted in Small/7 settings.}
\label{split}
\end{table}

\newpage

\section{Conclusion}
In this paper, we improve the spatial-shift MLP (S$^2$-MLP) and propose an S$^2$-MLPv2 model. It expands the feature map and splits the expanded feature map into three splits. It shifts each split individually and then fuses the split feature maps through split-attention. Meanwhile, we exploit the  hierarchical pyramid  to improve its capability of modeling fine-grained details for  higher recognition accuracy. Using $55$M parameters, our S$^2$-MLPv2-Medium model achieves $83.6\%$ top-1 accuracy on ImageNet1K dataset using $224\times 224$ images without external training datasets, which is the state-of-the-art performance among MLP-based methods. Meanwhile, compared with Transformer-based methods, our
S$^2$-MLPv2 model has achieved comparable accuracy without self-attention and fewer parameters.

Compared with the pioneering MLP-based works such as MLP-mixer, ResMLP as well as recent MLP-like models including Vision Permutator and GFNet, another important advantage of the spatial-shift MLP is that the shapes of spatial-shift MLPs are invariant to the input scale of images. Thus, the spatial-shift MLP model pre-trained by  images of a specific scale can be well adopted for  down-stream tasks with various-sized input images.

The future work will be devoted to continuously improving the image recognition accuracy of the spatial-shift MLP architecture. A promising and straightforward direction is to attempt smaller-size patches and the more advanced four-level pyramid as CycleMLP and AS-MLP  for further reducing the FLOPs and  shortening the recognition gap between the Transformer-based models.

\bibliography{egbib}

\begin{thebibliography}{41}
\providecommand{\natexlab}[1]{#1}
\providecommand{\url}[1]{\texttt{#1}}
\expandafter\ifx\csname urlstyle\endcsname\relax
  \providecommand{\doi}[1]{doi: #1}\else
  \providecommand{\doi}{doi: \begingroup \urlstyle{rm}\Url}\fi

\bibitem[Chen et~al.(2021{\natexlab{a}})Chen, Xie, Ge, Liang, and
  Luo]{chen2021cyclemlp}
Shoufa Chen, Enze Xie, Chongjian Ge, Ding Liang, and Ping Luo.
\newblock Cyclemlp: A mlp-like architecture for dense prediction.
\newblock \emph{arXiv preprint arXiv:2107.10224}, 2021{\natexlab{a}}.

\bibitem[Chen et~al.(2021{\natexlab{b}})Chen, Cheng, Gan, Yuan, Zhang, and
  Wang]{chen2021chasing}
Tianlong Chen, Yu~Cheng, Zhe Gan, Lu~Yuan, Lei Zhang, and Zhangyang Wang.
\newblock Chasing sparsity in vision transformers: An end-to-end exploration.
\newblock \emph{arXiv preprint arXiv:2106.04533}, 2021{\natexlab{b}}.

\bibitem[Chu et~al.(2021{\natexlab{a}})Chu, Tian, Wang, Zhang, Ren, Wei, Xia,
  and Shen]{chu2021twins}
Xiangxiang Chu, Zhi Tian, Yuqing Wang, Bo~Zhang, Haibing Ren, Xiaolin Wei,
  Huaxia Xia, and Chunhua Shen.
\newblock Twins: Revisiting the design of spatial attention in vision
  transformers.
\newblock \emph{arXiv preprint arXiv:2104.13840}, 2021{\natexlab{a}}.

\bibitem[Chu et~al.(2021{\natexlab{b}})Chu, Tian, Zhang, Wang, Wei, Xia, and
  Shen]{chu2021conditional}
Xiangxiang Chu, Zhi Tian, Bo~Zhang, Xinlong Wang, Xiaolin Wei, Huaxia Xia, and
  Chunhua Shen.
\newblock Conditional positional encodings for vision transformers.
\newblock \emph{arXiv preprint arXiv:2102.10882}, 2021{\natexlab{b}}.

\bibitem[Cubuk et~al.(2020)Cubuk, Zoph, Shlens, and Le]{cubuk2020randaugment}
Ekin~D Cubuk, Barret Zoph, Jonathon Shlens, and Quoc~V Le.
\newblock Randaugment: Practical automated data augmentation with a reduced
  search space.
\newblock In \emph{Proceedings of the IEEE/CVF Conference on Computer Vision
  and Pattern Recognition Workshops}, pp.\  702--703, 2020.

\bibitem[Deng et~al.(2009)Deng, Dong, Socher, Li, Li, and Li]{deng2009imagenet}
Jia Deng, Wei Dong, Richard Socher, Li{-}Jia Li, Kai Li, and Fei{-}Fei Li.
\newblock Imagenet: {A} large-scale hierarchical image database.
\newblock In \emph{Proceedings of the 2009 {IEEE} Computer Society Conference
  on Computer Vision and Pattern Recognition (CVPR)}, pp.\  248--255, Miami,
  FL, 2009.

\bibitem[Dong et~al.(2021)Dong, Bao, Chen, Zhang, Yu, Yuan, Chen, and
  Guo]{dong2021cswin}
Xiaoyi Dong, Jianmin Bao, Dongdong Chen, Weiming Zhang, Nenghai Yu, Lu~Yuan,
  Dong Chen, and Baining Guo.
\newblock Cswin transformer: A general vision transformer backbone with
  cross-shaped windows.
\newblock \emph{arXiv preprint arXiv:2107.00652}, 2021.

\bibitem[Dosovitskiy et~al.(2021)Dosovitskiy, Beyer, Kolesnikov, Weissenborn,
  Zhai, Unterthiner, Dehghani, Minderer, Heigold, Gelly, Uszkoreit, and
  Houlsby]{dosovitskiy2020image}
Alexey Dosovitskiy, Lucas Beyer, Alexander Kolesnikov, Dirk Weissenborn,
  Xiaohua Zhai, Thomas Unterthiner, Mostafa Dehghani, Matthias Minderer, Georg
  Heigold, Sylvain Gelly, Jakob Uszkoreit, and Neil Houlsby.
\newblock An image is worth 16x16 words: Transformers for image recognition at
  scale.
\newblock In \emph{Proceedings of the 9th International Conference on Learning
  Representations (ICLR)}, Virtual Event, 2021.

\bibitem[Fan et~al.(2021)Fan, Xiong, Mangalam, Li, Yan, Malik, and
  Feichtenhofer]{fan2021multiscale}
Haoqi Fan, Bo~Xiong, Karttikeya Mangalam, Yanghao Li, Zhicheng Yan, Jitendra
  Malik, and Christoph Feichtenhofer.
\newblock Multiscale vision transformers.
\newblock \emph{arXiv preprint arXiv:2104.11227}, 2021.

\bibitem[Guo et~al.(2021)Guo, Liu, Mu, and Hu]{guo2021beyond}
Meng-Hao Guo, Zheng-Ning Liu, Tai-Jiang Mu, and Shi-Min Hu.
\newblock Beyond self-attention: External attention using two linear layers for
  visual tasks.
\newblock \emph{arXiv preprint arXiv:2105.02358}, 2021.

\bibitem[Han et~al.(2021)Han, Xiao, Wu, Guo, Xu, and Wang]{han2021transformer}
Kai Han, An~Xiao, Enhua Wu, Jianyuan Guo, Chunjing Xu, and Yunhe Wang.
\newblock Transformer in transformer.
\newblock \emph{arXiv preprint arXiv:2103.00112}, 2021.

\bibitem[He et~al.(2016)He, Zhang, Ren, and Sun]{he2016deep}
Kaiming He, Xiangyu Zhang, Shaoqing Ren, and Jian Sun.
\newblock Deep residual learning for image recognition.
\newblock In \emph{Proceedings of the 2016 {IEEE} Conference on Computer Vision
  and Pattern Recognition (CVPR)}, pp.\  770--778, Las Vegas, NV, 2016.

\bibitem[Heo et~al.(2021)Heo, Yun, Han, Chun, Choe, and Oh]{heo2021pit}
Byeongho Heo, Sangdoo Yun, Dongyoon Han, Sanghyuk Chun, Junsuk Choe, and
  Seong~Joon Oh.
\newblock Rethinking spatial dimensions of vision transformers.
\newblock \emph{arXiv: 2103.16302}, 2021.

\bibitem[Hou et~al.(2021)Hou, Jiang, Yuan, Cheng, Yan, and Feng]{hou2021vision}
Qibin Hou, Zihang Jiang, Li~Yuan, Ming-Ming Cheng, Shuicheng Yan, and Jiashi
  Feng.
\newblock Vision permutator: A permutable mlp-like architecture for visual
  recognition.
\newblock \emph{arXiv preprint arXiv:2106.12368}, 2021.

\bibitem[Huang et~al.(2016)Huang, Sun, Liu, Sedra, and
  Weinberger]{huang2016deep}
Gao Huang, Yu~Sun, Zhuang Liu, Daniel Sedra, and Kilian~Q Weinberger.
\newblock Deep networks with stochastic depth.
\newblock In \emph{European conference on computer vision}, pp.\  646--661.
  Springer, 2016.

\bibitem[Huang et~al.(2021)Huang, Ben, Luo, Cheng, Yu, and
  Fu]{huang2021shuffle}
Zilong Huang, Youcheng Ben, Guozhong Luo, Pei Cheng, Gang Yu, and Bin Fu.
\newblock Shuffle transformer: Rethinking spatial shuffle for vision
  transformer.
\newblock \emph{arXiv preprint arXiv:2106.03650}, 2021.

\bibitem[Laine \& Aila(2016)Laine and Aila]{laine2016temporal}
Samuli Laine and Timo Aila.
\newblock Temporal ensembling for semi-supervised learning.
\newblock \emph{arXiv preprint arXiv:1610.02242}, 2016.

\bibitem[Lian et~al.(2021)Lian, Yu, Sun, and Gao]{lian2021mlp}
Dongze Lian, Zehao Yu, Xing Sun, and Shenghua Gao.
\newblock As-mlp: An axial shifted mlp architecture for vision.
\newblock \emph{arXiv preprint arXiv:2107.08391}, 2021.

\bibitem[Liu et~al.(2021{\natexlab{a}})Liu, Dai, So, and Le]{liu2021pay}
Hanxiao Liu, Zihang Dai, David~R So, and Quoc~V Le.
\newblock Pay attention to mlps.
\newblock \emph{arXiv preprint arXiv:2105.08050}, 2021{\natexlab{a}}.

\bibitem[Liu et~al.(2021{\natexlab{b}})Liu, Lin, Cao, Hu, Wei, Zhang, Lin, and
  Guo]{liu2021swin}
Ze~Liu, Yutong Lin, Yue Cao, Han Hu, Yixuan Wei, Zheng Zhang, Stephen Lin, and
  Baining Guo.
\newblock Swin transformer: Hierarchical vision transformer using shifted
  windows.
\newblock \emph{arXiv preprint arXiv:2103.14030}, 2021{\natexlab{b}}.

\bibitem[Loshchilov \& Hutter(2019)Loshchilov and
  Hutter]{loshchilov2018decoupled}
Ilya Loshchilov and Frank Hutter.
\newblock Decoupled weight decay regularization.
\newblock In \emph{Proceedings of the 7th International Conference on Learning
  Representations (ICLR)}, New Orleans, LA, 2019.

\bibitem[Radosavovic et~al.(2020)Radosavovic, Kosaraju, Girshick, He, and
  Doll{\'{a}}r]{radosavovic2020designing}
Ilija Radosavovic, Raj~Prateek Kosaraju, Ross~B. Girshick, Kaiming He, and
  Piotr Doll{\'{a}}r.
\newblock Designing network design spaces.
\newblock In \emph{Proceedings of the 2020 {IEEE/CVF} Conference on Computer
  Vision and Pattern Recognition (CVPR)}, pp.\  10425--10433, Seattle, WA,
  2020.

\bibitem[Rao et~al.(2021{\natexlab{a}})Rao, Zhao, Liu, Lu, Zhou, and
  Hsieh]{rao2021dynamicvit}
Yongming Rao, Wenliang Zhao, Benlin Liu, Jiwen Lu, Jie Zhou, and Cho-Jui Hsieh.
\newblock Dynamicvit: Efficient vision transformers with dynamic token
  sparsification.
\newblock \emph{arXiv preprint arXiv:2106.02034}, 2021{\natexlab{a}}.

\bibitem[Rao et~al.(2021{\natexlab{b}})Rao, Zhao, Zhu, Lu, and
  Zhou]{rao2021global}
Yongming Rao, Wenliang Zhao, Zheng Zhu, Jiwen Lu, and Jie Zhou.
\newblock Global filter networks for image classification.
\newblock \emph{arXiv preprint arXiv:2107.00645}, 2021{\natexlab{b}}.

\bibitem[Szegedy et~al.(2016)Szegedy, Vanhoucke, Ioffe, Shlens, and
  Wojna]{szegedy2016rethinking}
Christian Szegedy, Vincent Vanhoucke, Sergey Ioffe, Jonathon Shlens, and
  Zbigniew Wojna.
\newblock Rethinking the inception architecture for computer vision.
\newblock In \emph{Proceedings of the 2016 {IEEE} Conference on Computer Vision
  and Pattern Recognition (CVPR)}, pp.\  2818--2826, Las Vegas, NV, 2016.

\bibitem[Tan \& Le(2019)Tan and Le]{tan2019efficientnet}
Mingxing Tan and Quoc~V. Le.
\newblock Efficientnet: Rethinking model scaling for convolutional neural
  networks.
\newblock In \emph{Proceedings of the 36th International Conference on Machine
  Learning (ICML)}, pp.\  6105--6114, Long Beach, CA, 2019.

\bibitem[Tolstikhin et~al.(2021)Tolstikhin, Houlsby, Kolesnikov, Beyer, Zhai,
  Unterthiner, Yung, Steiner, Keysers, Uszkoreit, Lucic, and
  Dosovitskiy]{tolstikhin2021mlp}
Ilya Tolstikhin, Neil Houlsby, Alexander Kolesnikov, Lucas Beyer, Xiaohua Zhai,
  Thomas Unterthiner, Jessica Yung, Andreas Steiner, Daniel Keysers, Jakob
  Uszkoreit, Mario Lucic, and Alexey Dosovitskiy.
\newblock {MLP-Mixer}: An all-{MLP} architecture for vision.
\newblock \emph{arXiv preprint arXiv:2105.01601}, 2021.

\bibitem[Touvron et~al.(2020)Touvron, Cord, Douze, Massa, Sablayrolles, and
  J{\'e}gou]{touvron2020training}
Hugo Touvron, Matthieu Cord, Matthijs Douze, Francisco Massa, Alexandre
  Sablayrolles, and Herv{\'e} J{\'e}gou.
\newblock Training data-efficient image transformers \& distillation through
  attention.
\newblock \emph{arXiv preprint arXiv:2012.12877}, 2020.

\bibitem[Touvron et~al.(2021{\natexlab{a}})Touvron, Bojanowski, Caron, Cord,
  El-Nouby, Grave, Joulin, Synnaeve, Verbeek, and J{\'e}gou]{touvron2021resmlp}
Hugo Touvron, Piotr Bojanowski, Mathilde Caron, Matthieu Cord, Alaaeldin
  El-Nouby, Edouard Grave, Armand Joulin, Gabriel Synnaeve, Jakob Verbeek, and
  Herv{\'e} J{\'e}gou.
\newblock {ResMLP}: Feedforward networks for image classification with
  data-efficient training.
\newblock \emph{arXiv preprint arXiv:2105.03404}, 2021{\natexlab{a}}.

\bibitem[Touvron et~al.(2021{\natexlab{b}})Touvron, Cord, Sablayrolles,
  Synnaeve, and J{\'e}gou]{touvron2021going}
Hugo Touvron, Matthieu Cord, Alexandre Sablayrolles, Gabriel Synnaeve, and
  Herv{\'e} J{\'e}gou.
\newblock Going deeper with image transformers.
\newblock \emph{arXiv preprint arXiv:2103.17239}, 2021{\natexlab{b}}.

\bibitem[Wang et~al.(2021{\natexlab{a}})Wang, Xie, Li, Fan, Song, Liang, Lu,
  Luo, and Shao]{wang2021pvtv2}
Wenhai Wang, Enze Xie, Xiang Li, Deng-Ping Fan, Kaitao Song, Ding Liang, Tong
  Lu, Ping Luo, and Ling Shao.
\newblock Pvtv2: Improved baselines with pyramid vision transformer.
\newblock \emph{arXiv preprint arXiv:2106.13797}, 2021{\natexlab{a}}.

\bibitem[Wang et~al.(2021{\natexlab{b}})Wang, Xie, Li, Fan, Song, Liang, Lu,
  Luo, and Shao]{wang2021pyramid}
Wenhai Wang, Enze Xie, Xiang Li, Deng-Ping Fan, Kaitao Song, Ding Liang, Tong
  Lu, Ping Luo, and Ling Shao.
\newblock Pyramid vision transformer: A versatile backbone for dense prediction
  without convolutions.
\newblock \emph{arXiv preprint arXiv:2102.12122}, 2021{\natexlab{b}}.

\bibitem[Yang et~al.(2021)Yang, Li, Zhang, Dai, Xiao, Yuan, and
  Gao]{yang2021focal}
Jianwei Yang, Chunyuan Li, Pengchuan Zhang, Xiyang Dai, Bin Xiao, Lu~Yuan, and
  Jianfeng Gao.
\newblock Focal self-attention for local-global interactions in vision
  transformers.
\newblock \emph{arXiv preprint arXiv:2107.00641}, 2021.

\bibitem[Yu et~al.(2021{\natexlab{a}})Yu, Li, Cai, Sun, and
  Li]{yu2021rethinking}
Tan Yu, Xu~Li, Yunfeng Cai, Mingming Sun, and Ping Li.
\newblock Rethinking token-mixing mlp for mlp-based vision backbone.
\newblock \emph{arXiv preprint arXiv:2106.14882}, 2021{\natexlab{a}}.

\bibitem[Yu et~al.(2021{\natexlab{b}})Yu, Li, Cai, Sun, and Li]{yu2021s}
Tan Yu, Xu~Li, Yunfeng Cai, Mingming Sun, and Ping Li.
\newblock S2-{MLP}: Spatial-shift mlp architecture for vision.
\newblock \emph{arXiv preprint arXiv:2106.07477}, 2021{\natexlab{b}}.

\bibitem[Yuan et~al.(2021)Yuan, Chen, Wang, Yu, Shi, Jiang, Tay, Feng, and
  Yan]{yuan2021tokens}
Li~Yuan, Yunpeng Chen, Tao Wang, Weihao Yu, Yujun Shi, Zihang Jiang, Francis~EH
  Tay, Jiashi Feng, and Shuicheng Yan.
\newblock Tokens-to-token {ViT}: Training vision transformers from scratch on
  imagenet.
\newblock \emph{arXiv preprint arXiv:2101.11986}, 2021.

\bibitem[Yun et~al.(2019)Yun, Han, Chun, Oh, Yoo, and Choe]{yun2019cutmix}
Sangdoo Yun, Dongyoon Han, Sanghyuk Chun, Seong~Joon Oh, Youngjoon Yoo, and
  Junsuk Choe.
\newblock Cutmix: Regularization strategy to train strong classifiers with
  localizable features.
\newblock In \emph{Proceedings of the 2019 {IEEE/CVF} International Conference
  on Computer Vision (ICCV)}, pp.\  6022--6031, Seoul, Korea, 2019.

\bibitem[Zhang et~al.(2020)Zhang, Wu, Zhang, Zhu, Zhang, Lin, Sun, He, Muller,
  Manmatha, Li, and Smola]{zhang2020resnest}
Hang Zhang, Chongruo Wu, Zhongyue Zhang, Yi~Zhu, Zhi Zhang, Haibin Lin, Yue
  Sun, Tong He, Jonas Muller, R.~Manmatha, Mu~Li, and Alexander Smola.
\newblock Resnest: Split-attention networks.
\newblock \emph{arXiv preprint arXiv:2004.08955}, 2020.

\bibitem[Zhang et~al.(2018)Zhang, Ciss{\'{e}}, Dauphin, and
  Lopez{-}Paz]{zhang2017mixup}
Hongyi Zhang, Moustapha Ciss{\'{e}}, Yann~N. Dauphin, and David Lopez{-}Paz.
\newblock mixup: Beyond empirical risk minimization.
\newblock In \emph{Proceedings of the 6th International Conference on Learning
  Representations (ICLR)}, Vancouver, Canada, 2018.

\bibitem[Zhang et~al.(2021)Zhang, Zhang, Zhao, Chen, and
  Pfister]{zhang2021aggregating}
Zizhao Zhang, Han Zhang, Long Zhao, Ting Chen, and Tomas Pfister.
\newblock Aggregating nested transformers.
\newblock \emph{arXiv preprint arXiv:2105.12723}, 2021.

\bibitem[Zhong et~al.(2020)Zhong, Zheng, Kang, Li, and Yang]{zhong2020random}
Zhun Zhong, Liang Zheng, Guoliang Kang, Shaozi Li, and Yi~Yang.
\newblock Random erasing data augmentation.
\newblock In \emph{Proceedings of the Thirty-Fourth {AAAI} Conference on
  Artificial Intelligence (AAAI)}, pp.\  13001--13008, New York, NY, 2020.

\end{thebibliography}
\bibliographystyle{conference}

\end{document}